\pdfoutput=1

\documentclass[11pt]{article}

\usepackage[final]{emnlp2021}

\usepackage{times}
\usepackage{latexsym}
\usepackage[inline]{enumitem}

\usepackage{amssymb}
\usepackage{enumitem}

\usepackage{amsmath}
\usepackage{graphicx}
\usepackage{url}
\usepackage{xcolor}
\usepackage[colorinlistoftodos]{todonotes}
\usepackage[T1]{fontenc}

\usepackage[utf8]{inputenc}

\usepackage{microtype}
\usepackage{makecell}

\usepackage{fancyhdr}

\title{Pruned Graph Neural Network for Short Story Ordering}

\author{Melika Golestani \and Zeinab Borhanifard \and Farnaz Tahmasebian \and Heshaam Faili\\\AND School of Electrical and Computer Engineering, College of Engineering, University of Tehran, Tehran, Iran\\\texttt{\{melika.golestani,borhanifardz,hfaili\}@ut.ac.ir}\\\AND Alumni of Emory University, Atlanta, GA, USA\\\texttt{tahmasebian.farnaz@gmail.com}}

\begin{document}
\maketitle
\begin{abstract}
Text coherence is a fundamental problem in natural language generation and understanding. Organizing sentences into an order that maximizes coherence is known as sentence ordering. This paper is proposing a new approach based on the graph neural network approach to encode a set of sentences and learn orderings of short stories. We propose a new method for constructing sentence-entity graphs of short stories to create the edges between sentences and reduce noise in our graph by replacing the pronouns with their referring entities. We improve the sentence ordering by introducing an aggregation method based on majority voting of state-of-the-art methods and our proposed one. Our approach employs a BERT-based model to learn semantic representations of the sentences. 
The results demonstrate that the proposed method significantly outperforms existing baselines on a corpus of short stories with a new state-of-the-art performance in terms of Perfect Match Ratio (PMR) and Kendall's Tau ($\tau$) metrics. More precisely, our method increases PMR and $\tau$ criteria by more than 5\% and 4.3\%, respectively. These outcomes highlight the benefit of forming the edges between sentences based on their cosine similarity. We also observe that replacing pronouns with their referring entities effectively encodes sentences in sentence-entity graphs.

\end{abstract}

\section{Introduction}

Text coherence is a fundamental problem in natural language generation and understanding. A coherent text adheres to a logical order of events which facilitates better understanding. One of the subtasks in coherence modeling, called sentence ordering, refers to organizing shuffled sentences into an order that maximizes coherence \citep{barzilay2008modeling}. Several downstream applications benefit from this task to assemble sound and easy-to-understand texts, such as extraction-based multi-document summarization \citep{barzilay2002inferring,galanis2012extractive,nallapati2017summarunner,logeswaran2018sentence}, natural language generation \citep{reiter1997building}, retrieval-based question answering \citep{liu-etal-2018-stochastic,yu2018qanet}, concept-to-text generation \citep{konstas2012concept}, storytelling \citep{fan-etal-2019-strategies,hu2020makes,zhu-etal-2020-scriptwriter}, opinion generation \citep{yanase2015learning}, conversational analysis \citep{zeng2018microblog}, image captioning \citep{anderson2018bottom}, recipe generation \citep{chandu2019storyboarding}, and discourse coherence \citep{elsner2007unified,barzilay2008modeling,guinaudeau2013graph}.

In early studies, researchers modeled sentence structure using hand-crafted linguistic features \citep{lapata2003probabilistic,barzilay2004catching,elsner2007unified,barzilay2008modeling}, nonetheless, these features are domain-specific. Therefore recent studies employed deep learning techniques to solve sentence ordering tasks \citep{10.1007/978-3-030-83527-9_27,logeswaran2018sentence,gong2016end,li-jurafsky-2017-neural,chen2016neural}. 

\citep{cui2018deep} used graph neural networks, called ATTOrderNet, to accomplish this task. They used a self-attention mechanism combined with LSTMs to encode input sentences. Their method could get a reliable representation of the set of sentences regardless of their input order. In this representation, an ordered sequence is generated using a pointer network. Since ATTOrderNet is based on fully connected graph representations, it causes to build an association among some irrelevant sentences, which introduces massive noise into the network. Furthermore, since a self-attention mechanism only uses the information at the sentence level, other potentially helpful information such as entity information is missed. 

To overcome these drawbacks, \citet{ijcai2019-0748} developed the Sentence-Entity Graph (SE-Graph), which adds entities to the graph. While in the ATTOrderNet, every node is a sentence representation, SE-Graph consists of two types of nodes: sentence and entity\footnote{the entity should be common to at least two sentences}. Moreover, the edges come in two forms called SS and SE:

\begin{itemize}
    \item SS: this edge is between sentence nodes that share a common entity,
    \item SE: this edge connects a sentence and an entity within the sentence labeled with the entity's role.
\end{itemize}
\hbadness=99999  
However, the introduced methods perform poorly for short story reordering tasks. 
The SE-graph solution seems effective for long texts but not for short stories. In this paper, we suggest modifications to the introduced graph methods to improve the performance for short story reordering tasks. Some issues arise in short stories: First, entities are often not repeated in multiple sentences in a short story or text; instead, pronouns refer to an entity. To address this problem, we improve the semantic graph by replacing the pronouns with their corresponding entities. Another issue is a high correlation between the sentences in a short story along with a high commonality of entities across sentences, which leads us to end up with a complete graph in most cases. Our solution is moving towards a Pruned Graph (PG). 

As with the ATTOrderNet and SE-Graph networks, the PG architecture consists of three components: 
\begin{enumerate*}
    \item Sentence encoder based on SBERT-WK model \citep{wang2020sbert}, 
    \item Graph-based neural story encoder, and 
    \item Pointer network based decoder.
\end{enumerate*}

With PG, the network nodes and SE edges are created based on SE-Graph, in a way that the first and third components are the same. However, in the story encoding phase, after generating the nodes and SE edges based on SE-Graph method, the pruning phase is started on SS edges. This pruning process is defined as follow: each sentence edged out its neighbors with the first and second most cosine similarities
\citep{rahutomo2012semantic}. This method alleviates some problems of the previous two methods in the case of organizing short story sentences. It is noteworthy that pronouns are replaced with entities during pre-processing.

Finally, we present a method based on majority voting to combine our proposed graph network-based method with the state-of-art methods to benefit from each. Contributions of this study are as follows: \begin{enumerate}
    \item Proposing a new method based on graph networks to order sentences of a short stories corpus by:
    \begin{enumerate}
        \item suggesting a new method for creating the edges between sentences,
        \item creating a better sentence-entity graph for short stories by replacing pronouns in sentences with entities,
        \item Moreover, taking advantage of BERT-based sentence encoder.
    \end{enumerate}
    \item Using majority voting to combine sentence ordering methods.
\end{enumerate}

\section{Related Work}

\subsection{Sentence Ordering}

In early studies on sentence ordering, the structure of the document is modeled using hand-crafted linguistic features \citep{lapata2003probabilistic,barzilay2004catching,elsner2007unified,barzilay2008modeling}. \citet{lapata2003probabilistic} encoded sentences as vectors of linguistic features and used data to train a probabilistic transition model. \citet{barzilay2004catching} developed the content model in which topics in a specific domain are represented as states in an HMM. Some other like \citet{barzilay2008modeling} utilize the entity-based approach, which captures local coherence by modeling patterns of entity distributions. Other approaches used a combination of the entity grid and the content model \citet{elsner2007unified} or employed syntactic features \citep{louis2012coherence} in order to improve the model.

However, linguistic features are incredibly domain-specific, so applying these methods across different domains can decrease the performance. To overcome this limitation, recent works have used deep learning-based approaches. \citet{li2014model} proposes a neural model of distribution of sentence representations based on recurrent neural networks. In \citep{li-jurafsky-2017-neural}, graph-based neural models are used to generate a domain-independent neural model. \citet{agrawal-etal-2016-sort} introduced a method that involves combining two points elicited from the unary and pairwise model of sentences. \citet{chen2016neural} used an LSTM encoder and beam search to construct a pairwise model.
Based on a pointer network that provides advantages in capturing global coherence, \citet{gong2016end} developed an end-to-end approach that predicts order of sentences. In another work, by applying an encoder-decoder architecture based on LSTMs and attention mechanisms, \citet{logeswaran2018sentence} suggested a pairwise model and established the gold order by beam search. In \citep{pour2020new}, we presented a method that does not require any training corpus due to not having a training phase. We also developed a framework based on a sentence-level language model to solve the sentence ordering problem in \citep{10.1007/978-3-030-83527-9_27}. Moreover, in several other studies, including \citep{cui2018deep} and \citep{ijcai2019-0748}, graph neural networks are used to accomplish this task, as explained in the following.

\subsection{Graph Neural Networks in NLP}

Graph neural networks (GNN) have shown to be effective in NLP applications, including syntactic dependency trees \citep{marcheggiani-titov-2017-encoding}, neural machine translation \citep{beck-etal-2018-graph}, knowledge graphs \citep{wang-etal-2018-cross-lingual}, semantic graphs \cite{song-etal-2018-n}, sequence-to-graph learning \citep{Johnson2017LearningGS}, graph-to-sequence learning \citep{beck-etal-2018-graph}, sentence ordering \citep{ijcai2019-0748}, and multi-document summarization \citep{christensen-etal-2013-towards,yasunaga-etal-2017-graph}.

In particular, text classification is a common application of GNNs in natural language processing. A GNN infers document labels based on the relationships among documents or words \citep{10.5555/3294771.3294869}. \citet{christensen-etal-2013-towards} used a GNN in multi-document summarization. They create multi-document graphs which determine pairwise ordering constraints of sentences based on the discourse relationship between them. \citet{kipf2017semi} proposed Graph Convolutional Networks (GCN), which is used in \citet{yasunaga-etal-2017-graph} to generate sentence relation graphs. The final sentence embeddings indicate the graph representation and are utilized as inputs to achieve satisfactory results on multi-document summarization.

Another method is presented in \citet{marcheggiani-titov-2017-encoding} where a syntactic GCN is developed with a CNN/RNN as sentence encoder. The GCN indicates syntactic relations between words in a sentence. In a more recent work, \citet{ijcai2019-0748} proposed a graph-based neural network for sentence ordering, in which paragraphs are modeled as graphs where sentences and entities are the nodes. The method showed improvement in evaluation metrics for sentence ordering task. In this work, we explore the use of GRN for NLP tasks, especially to perform sentence-ordering on a corpus of short stories.

\section{Baselines}
\label{section:3}

This section introduces ATTOrderNet \citep{cui2018deep} and SE-Graph \citep{ijcai2019-0748}, which achieve state-of-the-art performances and serve as baseline for our work.

\subsection{ATTOrderNet}

ATTOrderNet introduced in \citep{cui2018deep} is a model using graph neural networks for sentence ordering. The model includes three components as follows: a sentence encoder based on Bi-LSTM, a paragraph encoder based on self-attention, and a pointer network-based decoder. In the sentence encoder, sentences are translated into distributional representations with a word embedding matrix. Then a sentence-level representation using the Bi-LSTM is learned. An average pooling layer follows multiple self-attention layers in the paragraph encoder. The paragraph encoder computes the attention scores for all pairs of sentences at different positions in the paragraph. Therefore, each sentence node is connected to all others where the encoder exploits latent dependency relations among sentences independent of their input order.

Having an input set of sentences, the decoder aims to predict a coherent order, identical to the original order. In this method, LSTM-based pointer networks are used to predict the correct sentence ordering from the final paragraph representation. 
Based on the sequence-to-sequence model, the pointer network-based decoders predict the correct sentence sequence \citep{10.5555/2969033.2969173}. Specifically, input tokens are encoded using the pointer network as summary vectors\footnote{The paragraph vector is nonetheless influenced by the permutations of input sentences.}, and the next token vector is decoded repeatedly. Finally, the output token sequence is derived from the output token vector.

\subsection{SE-Graph}
SE-Graph, similarly to ATTOrderNet, consists of three components: \begin{enumerate*}
    \item a sentence encoder based on Bi-LSTM,
    \item a paragraph encoder,
    \item a pointer network based decoder
\end{enumerate*}
Nevertheless, the difference between SE-Graph and ATTOrderNet is only in the encoder paragraph component, described in the following. In contrast to the fully connected graph representations explored by ATTOrderNet, \citet{ijcai2019-0748} represented input paragraphs as sentence-entity graphs. The SE-Graph includes two types of nodes: sentence and entity. The entity should be common to at least two sentences to be considered as a node of the graph. There are also two types of edges: SS edges that connect sentence nodes with at least a common entity, and SE edges that link a sentence with an entity within that and with a label of the entity's role. 
SE edges are labeled based on the syntactic role of the entity in the sentence, such as a subject, an object, or other. When an entity appears multiple times in a sentence with different roles, the role that has the highest rank is considered. The highest rank of roles is the subject role; after that are the object roles.
SE-Graph framework utilizes a GRN-based paragraph encoder that integrates the paragraph-level state along with the sentence-level state.

\section{Methodology}

In this section, first, the problem is formulated, second the dataset is introduced and explains why this dataset is suitable for the sentence ordering task. Then two methodologies are proposed. The first proposed method, called Pruned Graph, is based on graph networks, and the second is based on the majority voting to combine the outputs of three different models.

\subsection{Problem Formulation}

Consider $S(O)$ is a set of $n$ unordered sentences taken from a coherent text:

$O= {s_1, s_2, \dots, s_n}$,

\begin{equation}
s(o_1)> s(o_2) > \dots > s(o_n)
\end{equation}

The goal of sentence ordering is to find a permutation of sentences of O like $S(o\sp{\prime})$,

\begin{equation}
s({{o_1}\sp{\prime}}) > s({{o_2}\sp{\prime}}) > \ldots > s({{o_n}\sp{\prime}})
\end{equation}

that corresponds to the gold data arrangement. In other words, sentence ordering aims to restore the original orders:

\begin{equation}
s({o_1}^*) > s({o_2}^*) > \ldots > s({o_n}^*)
\end{equation}

Where ${S({o^*})}$ represents the original or gold order. As a result a correct output leads to ${S({o\sp{\prime}}) = S({o^*})}$. Based on the above definition and notions we propose our sentence ordering method.

\subsection{Dataset}

In this paper, we used a corpus of short stories, called ROCStories \citep{mostafazadeh-etal-2016-corpus}. It contains 98,162 commonsense stories, each with exactly five sentences and an average word count of 50. \citet{mostafazadeh-etal-2017-lsdsem} created ROCStories corpus for a shared task called LSDSem, in which models are supposed to predict the correct ending to short stories. 3,742 of the stories have two options for the final sentence. It is worth noting that humans generated all of the stories and options. 

We can learn sequences of daily events from this dataset because it contains some essential characteristics: The stories are \textbf{rich with causal and temporal relations among events}, which makes this dataset a highly useful resource for learning narrative structure across a wide range of events.
The dataset consists of a \textbf{comprehensive collection of daily and non-fictional short stories} useful for modeling the coherence of a text \citep{mostafazadeh-etal-2016-corpus}.

Due to these features, ROCStories can be used to learn sequences of sentences. Thus, the corpus is useful for organizing sentences in a text.

\subsection{Pruned Graph Sentence Ordering (PG)}

We propose a neural network based on the pruned graph for arranging the sentences of short stories, a modified version of the ATTOrderNet \citep{cui2018deep} and Sentence-Entity Graph \citep{ijcai2019-0748}. The PG method consists of three components: \textbf{sentence encoder}, \textbf{story encoder}, and \textbf{decoder}. In order to be a fair comparison, we used the same decoder as ATTOrderNet and SE-Graph. Due to space limitations, here we explain our sentence encoder and our story encoder. The Sentence encoder uses BERT encoding to encode sentences, while Story encoder uses a graph neural network for encoding stories. 

\subsubsection{Sentence Encoder: SBERT-WK}

We use fine-tuned pre-trained SBERT-WK model to encode sentences. BERT contains several layers, each of which captures a different linguistic characteristic. SBERT-WK found better sentence representations by fusing information from different layers, \citet{wang2020sbert}. The system geometrically analyzes space using a deep contextual model that is trained on both word-level and sentence-level, without further training. For each word in a sentence, it determines a unified word representation then computes the final sentence embedding vector based on the weighted average based on the word importance of the word representations. Even with a small embedding size of 768, SBERT-WK outperforms other methods by a significant margin on textual similarity tasks \citep{wang2020sbert}.

\subsubsection{Story Encoder}

To use graph neural networks for encoding stories, input stories should be represented as graphs. We propose a pruned graph (PG) representation instead of SE-Graph \citep{ijcai2019-0748} for encoding short stories. Nodes in PG are composed of sentences and entities. We replace pronouns with the entities they refer to since entities are not often repeated from one sentence to another during a short story\footnote{we use the Stanford's tool \citep{lee2011stanford}}, we will go into more detail in the experiments. We consider all nouns of an input story as entities at first. After that, we eliminate entities that do not occur more than once in the story. 

We can formalize our undirected pruned graphs as G = ($V_s$, $V_e$, E), where $V_s$ indicates the sentence-level nodes, $V_e$ denotes the entity-level nodes, and E represents edges. 
Edges in PG graphs are divided into two types: SS and SE. The SS type links two sentences in a story that have the highest or second-highest value of cosine similarity with each other; and the SE type links a sentence with an entity within that with a label of the entity's role. Equation~\ref{eq:1} shows the formula for calculating the cosine similarity, where CosSim is cosine similarity and $Emb_{s_i}$ represents vector of sentence i.

\begin{equation}
     CosSim (Emb_{s_i} , Emb_{s_j}) = \frac {Emb_{s_i} * Emb_{s_j}} {||Emb_{s_i}|| ||Emb_{s_j}||}
    \label{eq:1}
\end{equation}

SE edges are labeled according to the syntactic role of the entity in the sentence, such as a subject, an object, or other. The role that has the highest rank in an instance of an entity appearing multiple times is considered. The ranking is as follows: subject role, object roles, and other. The use of referring entities rather than pronouns is crucial.

Thus, sentence nodes are linked to both sentence and entity nodes, whereas an entity node is not connected to any other entity nodes. For graph encoding, we use GRN \citep{zhang-etal-2018-sentence}, which has been found effective for various kinds of graph encoding tasks. GRN used in our PG is the same as GRN in \citep{ijcai2019-0748}, so we do not explain it.

\subsection{Majority Voting}

 We combine the output of three methods to achieve better results in majority voting. Since the stories in Rocstories all have five sentences, there are 20 possible pair sentence orderings as follow:

\begin{enumerate*}

    \item $s_1 s_2$ or \item $s_2 s_1$, \item $s_1 s_3$ or \item $s_3 s_1$, \item $s_1 s_4$ or \item $s_4 s_1$, \item $s_1 s_5$ or \item $s_5 s_1$,
    \item $s_2 s_3$ or \item $s_3 s_2$, \item $s_2 s_4$ or \item $s_4 s_2$, \item $s_2 s_5$ or \item $s_5 s_2$,
    \item $s_3 s_4$ or \item $s_4 s_3$, \item $s_3 s_5$ or \item $s_5 s_3$,
    \item $s_4 s_5$ or \item $s_5 s_4$
\end{enumerate*}

Each suggested order for a story includes 10 of the above pair orderings, either of the two pair orderings that have an "or" between them. Through majority voting\footnote{For example, either $s_1 s_2$ or $s_2 s_1$ occurs, and without a doubt, the co-occurrence of these is a vast and impossible contradiction.}, we can combine the outputs of three separate methods to generate a final order.

According to the number of occurrences in each of the three output arrangements, we assign scores to each of the 20 possible pairings. As a result, each of these possible pairings is scored between 0 and 3. 0 indicates that this pairing does not appear in any of the three methods' outputs, while 3 indicates that it appears in all of them. In the end, all pairs with a greater score of 1 occur in the final orderings\footnote{Suppose the outputs of the three methods for arranging $sentence_1$ ($s_1$) and $sentence_2$ ($s_2$) are: Method 1: $s_1$$s_2$, Method 2: $s_1$$s_2$, and Method 3: $s_2$$s_1$. Therefore, the order $s_1$$s_2$ gets two points and the order $s_2$$s_1$ gets one, so $s_1$$s_2$ applies to the final output.}. Indeed, these are ten pairs\footnote{either of the two pair orderings that have an "or" between them.}, and with the chosen pairs, the sentences of the story are arranged uniquely.

In the following subsection, we are proving that majority voting is a valid way to combine the outputs generated from three different methods for arranging sentences.
By using contradiction, we demonstrate the validity of the majority voting method for combining three distinct methods of sentence ordering to arrange two sentences. 

Assuming the majority voting of three methods fails to create an unique order, then two orders are possible, $s_1 s_2$, and $s_2 s_1$. In the first case, $s_1$ appears before $s_2$ in two or more outputs of the methods, and in the second case, $s_2$ appears before $s_1$ in two or more outputs. Due to the three methods, this assumption causes a contradiction. To ordering more than two sentences, it can be proved by induction.

\section{Experiment}

\subsection{Evaluation Metrics}

We use two standard metrics to evaluate the proposed model outputs that are commonly used in previous work: Kendall's tau and perfect match ratio, as described below.

\begin{itemize}

    \item \textbf{Kendall's Tau ($\tau$)}

Kendal's Tau \citep{lapata-2006-automatic} measures the quality of the output's ordering, computed as follows: 

\begin{equation}
\tau = 1-
\frac{(2*\#\: of\: Inversions)}
    {N*(N-1)/2} 
\label{eq:2}
\end{equation}

Where \(N\) represents the sequence length (i.e. the number of sentences of a story, which is always equal to 5 for ROCStories), and the inversions return the number of exchanges of the predicted order with the gold order for reconstructing the correct order. $\tau$ is always between -1 and 1, where the upper bound indicates that the predicted order is exactly the same as the gold order. This metric correlates reliably with human judgments, according to \citet{lapata-2006-automatic}.

    \item \textbf{Perfect Match Ratio (PMR)}

According to this ratio, each story is considered as a single unit, and a ratio of the number of correct orders is calculated. Therefore no penalties are given for incorrect permutations \citep{gong2016end}. PMR is formulated mathematically as follows:

\begin{equation}
PMR= {{1/N} {\sum_{1}^{N} {o_i\sp{\prime}} = {o_i^*}}}
\label{eq:3}
\end{equation}

where \(o_i\sp{\prime}\) represents the output order and \(o_i^*\) indicates the gold order. \(N\) specifies the sequence length. Since the length of all the stories of ROCStories is equal to 5, \(N\) in this study is always equal to 5. 
PMR values range from 0 to 1, with a higher value indicating better performance.

\end{itemize}

\subsection{Contrast Models}

We compare our PG to the state of the arts, namely the following: 
\begin{enumerate*}
    \item LSTM + PtrNet \citep{gong2016end},
    \item LSTM + Set2Seq \citep{logeswaran2018sentence},
    \item ATTOrderNet \citep{cui2018deep},
    \item SE-Graph \citep{ijcai2019-0748},
    \item HAN \citep{wang2019hierarchical},
    \item SLM \citep{10.1007/978-3-030-83527-9_27},
    \item RankTxNet ListMLE \citep{kumar2020deep},
    \item Enhancing PtrNet + Pairwise \citep{yin2020enhancing},
    \item B-TSort \citep{prabhumoye-etal-2020-topological}.
\end{enumerate*} 
We teach LSTM + PtrNet, ATTOrderNet, SE-Graph, and B-TSort on the ROCStories. The following is a brief description of the mentioned methods, but We refer to Section \ref{section:3} to explain ATTOrderNet and SE-Graph.

\citet{gong2016end} proposes LSTM + PtrNet as a method for ordering sentences. In this end-to-end method, pointer networks sort encrypted sentences after decoding them by LSTM. \citet{logeswaran2018sentence} recommended LSTM + Set2Seq. Their method encodes sentences, learns context representation by LSTM and attention mechanisms, and utilizes a pointer network-based decoder to predict sentences' order. A transformer followed by an LSTM was added to the sentence encoder in \citep{wang2019hierarchical} to capture word clues and dependencies between sentences; and so on, HAN is developed.

In \citep{10.1007/978-3-030-83527-9_27}, we developed the Sentence-level Language Model (SLM) for Sentence Ordering, consisting of a Sentence Encoder, a Story Encoder, and a Sentence Organizer. The sentence encoder encodes sentences into a vector using a fine-tuned pre-trained BERT. Hence, the embedding pays more attention to the sentence's crucial parts. Afterward, the story encoder uses a decoder-encoder architecture to learn the sentence-level language model. The learned vector from the hidden state is decoded, and this decoded vector is utilized to indicate the following sentence's candidate. Finally, the sentence organizer uses the cosine similarity as the scoring function in order to sort the sentences.

An attention-based ranking framework is presented in \citep{kumar2020deep} to address the task. The model uses a bidirectional sentence encoder and a self-attention-based transformer network to endcode paragraphs.
In \citep{yin2020enhancing}, an enhancing pointer network based on two pairwise ordering prediction modules, The FUTURE and HISTORY module, is employed to decode paragraphs. Based on the candidate sentence, the FUTURE module predicts the relative positions of other unordered sentences. Although, the HISTORY module determines the coherency between the candidate and previously ordered sentences.
And lastly, \citet{prabhumoye-etal-2020-topological} designed B-TSort, a pairwise ordering method, which is the current state-of-the-art method for sentence ordering. This method benefits from BERT and graph-based networks. Based on the relative pairwise ordering, graphs are constructed. Finally, the global order is derived by a topological sort algorithm on the graph.

\subsection{Setting}

For a fair comparison, we follow \citet{ijcai2019-0748}'s settings. Nevertheless, we use SBERT-WK's 768- dimension vectors for sentence embedding. Furthermore, the state sizes for sentence nodes are set to 768 in the GRN; The Batch size is 32. In preprocessing, we use Stanford's tool \citep{lee2011stanford} to replace pronouns with the referring entities.

\subsection{Results}

In this paper, we propose a new method based on graph networks for sentences ordering short stories called Pruned Graph (PG). In order to achieve this, we propose a new method for creating edges between sentences (by calculating the cosine similarity between sentences), and we create a better sentence-entity graph for short stories by replacing pronouns with the relevant entities. Besides, to make a better comparison, we also teach the following cases:

\begin{enumerate}
    \item All nodes in the graph are of the sentence type, and the graph is fully connected. In other words, we train ATTOrderNet on ROCStories\footnote{\cite{cui2018deep} did not train ATTOrderNet on the ROCStories dataset.}.
    \item The nodes include sentence and entity nodes, and each sentence's node has the edge over all other sentences' nodes (semi fully connected SE-Graph\footnote{Entity nodes are not connected to all nodes.}).
    \item The network comprises sentence and entity nodes, and every two sentences with at least one entity in common are connected (SE-Graph\footnote{We train SE-Graph on ROCStories since \cite{ijcai2019-0748} did not.}).
    \item Replacing pronouns with the relevant entities in SE-Graph (SE-Graph + Co-referencing).
    \item Similar to PG, but each sentence is connected to a sentence with the highest cosine similarity (semi $PG_1$). 
    \item Similar to PG; however, each sentence is connected to three other sentences based on their cosine similarity (semi $PG_3$).
    \item Pruned Graph with a Bi-LSTM based sentence encoder\footnote{To demonstrate the advantages of the PG's BERT-based sentence encoder, this component is considered exactly like the sentence encoder of SE-Graph and ATTOrderNet.}($PG\sp{\prime}$).
\end{enumerate}

Note that in the above methods, where the graph also contains the nodes of the entity, there is an edge between a sentence and an entity within it\footnote{Entity nodes can only have a link to sentence nodes.}. Table 1 reports the results of Pruned Graph (PG) and the above seven methods.

To get the training, validation, and testing datasets, we randomly split ROCStories into 8:1:1. Therefore, the training set includes 78,529 stories, the validation set contains 9,816 stories, and the testing set consists of 9,817 stories.

As shown in table \ref{table:1}, our PG beats all seven other methods. The results show that all three of our innovations to the graph-based method have improved the performance. Based on our analysis, the SBERT-WK sentence encoder is more beneficial than the Bi-LSTM. Our experiences also find that using referring entities instead of pronouns is helpful to create a more effective sentence-entity graph. Additionally, it indicates connecting each sentence to two others using cosine similarity is efficient to encode a story.

\begin{table}[ht]
 \begin{center}
  \begin{tabular}{||c |c |c||} 
  \hline
  \textbf{Method} & \textbf{$\tau$} & \textbf{PMR} \\ [0.5ex]
  \hline
  ATTOrderNet & 0.7364 & 0.4030 \\ \hline
  \makecell{Fully connected \\ SE-Graph} & 0.7300 & 0.3927 \\
  \hline
  SE-Graph & 0.7133 & 0.3687 \\
  \hline
  \makecell{SE-Graph + \\ Co-ref} & 0.7301 & 0.3981\\
  \hline
  PG+ 1 SS & 0.7534 & 0.4349 \\
  \hline
  PG + 3 SS & 0.7379 & 0.4100 \\
  \hline
  \makecell{PG + Bi-LSTM-based \\ sentence encoder} & 0.7852 & 0.4769 \\
  \hline
  Pruned Graph & 0.8220 & 0.5373 \\[0.5ex] 
  \hline
  \end{tabular}
  \caption{Reporting the results of the proposed network called Pruned Graph in comparison with the seven methods mentioned above}
  \label{table:1}
 \end{center}
 \end{table}

Table \ref{table:2} reports the results of the proposed method of this paper in comparison with competitors. When compared with ATTOrdeNet, PG improved the Tau by over 8.5\% as well as PMR by 13.5\%. Furthermore, the Tau is increased by 10.8\% and the PMR by more than 16.8\% compared to SE-Graph. PG outperforms the state-of-the-art on ROCStories with a more tthan 1.8\% increase in pmr and a more than 3.9\% improvement in $\tau$.

\begin{table}[ht]
 \begin{center}
  \begin{tabular}{||c| c| c||} 
  \hline
  \textbf{Model} & \textbf{$\tau$} & \textbf{PMR} \\ [0.5ex]
  \hline
  LSTM+PtrNet & 0.7230 & 0.3647 \\
  \hline
  LSTM+Set2Seq & 0.7112 & 0.3581 \\
  \hline
  ATTOrderNet & 0.7364 & 0.4030 \\ 
  \hline
  SE-Graph & 0.7133 & 0.3687 \\
  \hline
  HAN & 0.7322 & 0.3962 \\
  \hline
  SLM & 0.7547 & 0.4064 \\
  \hline
  RankTxNet ListMLE & 0.7602 & 0.3602 \\
  \hline
  \makecell{Enhancing PtrNet + \\ Pairwise } & 0.7681 & 0.4600 \\
  \hline
  B-TSort & 0.8039 & 0.4980 \\
  \hline
  Our Pruned Graph & 0.8220 & 0.5373 \\[1ex] 
  \hline
  \end{tabular}
  \caption{Results of our PG compared to baselines and competitors}
  \label{table:2}
 \end{center}
\end{table}

Finally, we merged the outputs of the three methods using the majority voting method, including Enhancing PtrNet, B-TSort, and Our Pruned Graph. Table \ref{table:3} shows the results of the combination, which improves the PMR and $\tau$ criteria by more than 5\% and 4.3\% on ROCStories, respectively.

\begin{table}[ht]
 \begin{center}
  \begin{tabular}{||c| c |c||} 
  \hline
  \textbf{Method} & \textbf{$\tau$} & \textbf{PMR} \\ [0.5ex]
  \hline
  combination & 0.8470 & 0.5488 \\ [0.5ex] 
  \hline
  \end{tabular}
  \caption{Results of combining of Enhancing PtrNet, B-TSort, and Pruned Graph using majority voting}
  \label{table:3}
 \end{center}
 \end{table}

\section{conclusion}

This paper introduced a graph-based neural framework to solve the sentence ordering task. This framework takes a set of randomly ordered sentences and outputs a coherent order of the sentences. The results demonstrate that SBERT-WK is a reliable model to encode sentences. Our analysis examined how the method is affected by using a Bi-LSTM model in the sentence encoder component. In addition, we found that replacing pronouns with their referring entities supplies a more informative sentence-entity graph to encode a story.
The experimental results indicate that our proposed graph-based neural model significantly outperforms on ROCStories dataset. Furthermore, we recommend a method for combining different methods of sentence ordering based on majority voting that achieves state-of-the-art performance in PMR and $\tau$ scores.
In future, we plan to apply the trained model on sentence ordering task to tackle other tasks including text generation, dialogue generation, text completion, retrieval-based QA, and extractive text summarization.

\bibliography{emnlp2021}

\begin{thebibliography}{51}
\expandafter\ifx\csname natexlab\endcsname\relax\def\natexlab#1{#1}\fi

\bibitem[{Agrawal et~al.(2016)Agrawal, Chandrasekaran, Batra, Parikh, and
  Bansal}]{agrawal-etal-2016-sort}
Harsh Agrawal, Arjun Chandrasekaran, Dhruv Batra, Devi Parikh, and Mohit
  Bansal. 2016.
\newblock \href {https://doi.org/10.18653/v1/D16-1091} {Sort story: Sorting
  jumbled images and captions into stories}.
\newblock In \emph{Proceedings of the 2016 Conference on Empirical Methods in
  Natural Language Processing}, pages 925--931, Austin, Texas. Association for
  Computational Linguistics.

\bibitem[{Anderson et~al.(2018)Anderson, He, Buehler, Teney, Johnson, Gould,
  and Zhang}]{anderson2018bottom}
Peter Anderson, Xiaodong He, Chris Buehler, Damien Teney, Mark Johnson, Stephen
  Gould, and Lei Zhang. 2018.
\newblock Bottom-up and top-down attention for image captioning and visual
  question answering.
\newblock In \emph{Proceedings of the IEEE conference on computer vision and
  pattern recognition}, pages 6077--6086.

\bibitem[{Barzilay and Elhadad(2002)}]{barzilay2002inferring}
Regina Barzilay and Noemie Elhadad. 2002.
\newblock Inferring strategies for sentence ordering in multidocument news
  summarization.
\newblock \emph{Journal of Artificial Intelligence Research}, 17:35--55.

\bibitem[{Barzilay and Lapata(2008)}]{barzilay2008modeling}
Regina Barzilay and Mirella Lapata. 2008.
\newblock Modeling local coherence: An entity-based approach.
\newblock \emph{Computational Linguistics}, 34(1):1--34.

\bibitem[{Barzilay and Lee(2004)}]{barzilay2004catching}
Regina Barzilay and Lillian Lee. 2004.
\newblock Catching the drift: Probabilistic content models, with applications
  to generation and summarization.
\newblock In \emph{Proceedings of the Human Language Technology Conference of
  the North American Chapter of the Association for Computational Linguistics:
  HLT-NAACL 2004}, pages 113--120.

\bibitem[{Beck et~al.(2018)Beck, Haffari, and Cohn}]{beck-etal-2018-graph}
Daniel Beck, Gholamreza Haffari, and Trevor Cohn. 2018.
\newblock \href {https://doi.org/10.18653/v1/P18-1026} {Graph-to-sequence
  learning using gated graph neural networks}.
\newblock In \emph{Proceedings of the 56th Annual Meeting of the Association
  for Computational Linguistics (Volume 1: Long Papers)}, pages 273--283,
  Melbourne, Australia. Association for Computational Linguistics.

\bibitem[{Chandu et~al.(2019)Chandu, Nyberg, and
  Black}]{chandu2019storyboarding}
Khyathi Chandu, Eric Nyberg, and Alan~W Black. 2019.
\newblock Storyboarding of recipes: grounded contextual generation.
\newblock In \emph{Proceedings of the 57th Annual Meeting of the Association
  for Computational Linguistics}, pages 6040--6046.

\bibitem[{Chen et~al.(2016)Chen, Qiu, and Huang}]{chen2016neural}
Xinchi Chen, Xipeng Qiu, and Xuanjing Huang. 2016.
\newblock Neural sentence ordering.
\newblock \emph{arXiv preprint arXiv:1607.06952}.

\bibitem[{Christensen et~al.(2013)Christensen, {Mausam}, Soderland, and
  Etzioni}]{christensen-etal-2013-towards}
Janara Christensen, {Mausam}, Stephen Soderland, and Oren Etzioni. 2013.
\newblock \href {https://www.aclweb.org/anthology/N13-1136} {Towards coherent
  multi-document summarization}.
\newblock In \emph{Proceedings of the 2013 Conference of the North {A}merican
  Chapter of the Association for Computational Linguistics: Human Language
  Technologies}, pages 1163--1173, Atlanta, Georgia. Association for
  Computational Linguistics.

\bibitem[{Cui et~al.(2018)Cui, Li, Chen, and Zhang}]{cui2018deep}
Baiyun Cui, Yingming Li, Ming Chen, and Zhongfei Zhang. 2018.
\newblock Deep attentive sentence ordering network.
\newblock In \emph{Proceedings of the 2018 Conference on Empirical Methods in
  Natural Language Processinga}, pages 4340--4349.

\bibitem[{Elsner et~al.(2007)Elsner, Austerweil, and
  Charniak}]{elsner2007unified}
Micha Elsner, Joseph Austerweil, and Eugene Charniak. 2007.
\newblock A unified local and global model for discourse coherence.
\newblock In \emph{Human Language Technologies 2007: The Conference of the
  North American Chapter of the Association for Computational Linguistics;
  Proceedings of the Main Conference}, pages 436--443.

\bibitem[{Fan et~al.(2019)Fan, Lewis, and Dauphin}]{fan-etal-2019-strategies}
Angela Fan, Mike Lewis, and Yann Dauphin. 2019.
\newblock \href {https://doi.org/10.18653/v1/P19-1254} {Strategies for
  structuring story generation}.
\newblock In \emph{Proceedings of the 57th Annual Meeting of the Association
  for Computational Linguistics}, pages 2650--2660, Florence, Italy.
  Association for Computational Linguistics.

\bibitem[{Galanis et~al.(2012)Galanis, Lampouras, and
  Androutsopoulos}]{galanis2012extractive}
Dimitrios Galanis, Gerasimos Lampouras, and Ion Androutsopoulos. 2012.
\newblock Extractive multi-document summarization with integer linear
  programming and support vector regression.
\newblock In \emph{Proceedings of COLING 2012}, pages 911--926.

\bibitem[{Golestani et~al.(2021)Golestani, Razavi, Borhanifard, Tahmasebian,
  and Faili}]{10.1007/978-3-030-83527-9_27}
Melika Golestani, Seyedeh~Zahra Razavi, Zeinab Borhanifard, Farnaz Tahmasebian,
  and Hesham Faili. 2021.
\newblock Using bert encoding and sentence-level language model for sentence
  ordering.
\newblock In \emph{Text, Speech, and Dialogue}, pages 318--330, Cham. Springer
  International Publishing.

\bibitem[{Gong et~al.(2016)Gong, Chen, Qiu, and Huang}]{gong2016end}
Jingjing Gong, Xinchi Chen, Xipeng Qiu, and Xuanjing Huang. 2016.
\newblock End-to-end neural sentence ordering using pointer network.
\newblock \emph{arXiv preprint arXiv:1611.04953}.

\bibitem[{Guinaudeau and Strube(2013)}]{guinaudeau2013graph}
Camille Guinaudeau and Michael Strube. 2013.
\newblock Graph-based local coherence modeling.
\newblock In \emph{Proceedings of the 51st Annual Meeting of the Association
  for Computational Linguistics (Volume 1: Long Papers)}, pages 93--103.

\bibitem[{Hamilton et~al.(2017)Hamilton, Ying, and
  Leskovec}]{10.5555/3294771.3294869}
William~L. Hamilton, Rex Ying, and Jure Leskovec. 2017.
\newblock Inductive representation learning on large graphs.
\newblock In \emph{Proceedings of the 31st International Conference on Neural
  Information Processing Systems}, NIPS'17, page 1025–1035, Red Hook, NY,
  USA. Curran Associates Inc.

\bibitem[{Hu et~al.(2020)Hu, Cheng, Gan, Liu, Gao, and Neubig}]{hu2020makes}
Junjie Hu, Yu~Cheng, Zhe Gan, Jingjing Liu, Jianfeng Gao, and Graham Neubig.
  2020.
\newblock What makes a good story? designing composite rewards for visual
  storytelling.
\newblock In \emph{Proceedings of the AAAI Conference on Artificial
  Intelligence}, volume~34, pages 7969--7976.

\bibitem[{Johnson(2017)}]{Johnson2017LearningGS}
D.~Johnson. 2017.
\newblock Learning graphical state transitions.
\newblock In \emph{ICLR}.

\bibitem[{Kipf and Welling(2017)}]{kipf2017semi}
Thomas~N Kipf and Max Welling. 2017.
\newblock Semi-supervised classification with graph convolutional networks.
\newblock \emph{arXiv preprint arXiv:1609.02907}.

\bibitem[{Konstas and Lapata(2012)}]{konstas2012concept}
Ioannis Konstas and Mirella Lapata. 2012.
\newblock Concept-to-text generation via discriminative reranking.
\newblock In \emph{Proceedings of the 50th Annual Meeting of the Association
  for Computational Linguistics (Volume 1: Long Papers)}, pages 369--378.

\bibitem[{Kumar et~al.(2020)Kumar, Brahma, Karnick, and Rai}]{kumar2020deep}
Pawan Kumar, Dhanajit Brahma, Harish Karnick, and Piyush Rai. 2020.
\newblock Deep attentive ranking networks for learning to order sentences.
\newblock In \emph{Proceedings of the AAAI Conference on Artificial
  Intelligence}, volume~34, pages 8115--8122.

\bibitem[{Lapata(2003)}]{lapata2003probabilistic}
Mirella Lapata. 2003.
\newblock Probabilistic text structuring: Experiments with sentence ordering.
\newblock In \emph{Proceedings of the 41st Annual Meeting of the Association
  for Computational Linguistics}, pages 545--552.

\bibitem[{Lapata(2006)}]{lapata-2006-automatic}
Mirella Lapata. 2006.
\newblock \href {https://doi.org/10.1162/coli.2006.32.4.471} {Automatic
  evaluation of information ordering: Kendall{'}s tau}.
\newblock \emph{Computational Linguistics}, 32(4):471--484.

\bibitem[{Lee et~al.(2011)Lee, Peirsman, Chang, Chambers, Surdeanu, and
  Jurafsky}]{lee2011stanford}
Heeyoung Lee, Yves Peirsman, Angel Chang, Nathanael Chambers, Mihai Surdeanu,
  and Dan Jurafsky. 2011.
\newblock Stanford’s multi-pass sieve coreference resolution system at the
  conll-2011 shared task.
\newblock In \emph{Proceedings of the 15th conference on computational natural
  language learning: Shared task}, pages 28--34. Association for Computational
  Linguistics.

\bibitem[{Li and Hovy(2014)}]{li2014model}
Jiwei Li and Eduard Hovy. 2014.
\newblock A model of coherence based on distributed sentence representation.
\newblock In \emph{Proceedings of the 2014 Conference on Empirical Methods in
  Natural Language Processing (EMNLP)}, pages 2039--2048.

\bibitem[{Li and Jurafsky(2017)}]{li-jurafsky-2017-neural}
Jiwei Li and Dan Jurafsky. 2017.
\newblock \href {https://doi.org/10.18653/v1/D17-1019} {Neural net models of
  open-domain discourse coherence}.
\newblock In \emph{Proceedings of the 2017 Conference on Empirical Methods in
  Natural Language Processing}, pages 198--209, Copenhagen, Denmark.
  Association for Computational Linguistics.

\bibitem[{Liu et~al.(2018)Liu, Shen, Duh, and Gao}]{liu-etal-2018-stochastic}
Xiaodong Liu, Yelong Shen, Kevin Duh, and Jianfeng Gao. 2018.
\newblock \href {https://doi.org/10.18653/v1/P18-1157} {Stochastic answer
  networks for machine reading comprehension}.
\newblock In \emph{Proceedings of the 56th Annual Meeting of the Association
  for Computational Linguistics (Volume 1: Long Papers)}, pages 1694--1704,
  Melbourne, Australia. Association for Computational Linguistics.

\bibitem[{Logeswaran et~al.(2018)Logeswaran, Lee, and
  Radev}]{logeswaran2018sentence}
Lajanugen Logeswaran, Honglak Lee, and Dragomir Radev. 2018.
\newblock Sentence ordering and coherence modeling using recurrent neural
  networks.
\newblock In \emph{Proceedings of the AAAI Conference on Artificial
  Intelligence}, volume~32.

\bibitem[{Louis and Nenkova(2012)}]{louis2012coherence}
Annie Louis and Ani Nenkova. 2012.
\newblock A coherence model based on syntactic patterns.
\newblock In \emph{Proceedings of the 2012 Joint Conference on Empirical
  Methods in Natural Language Processing and Computational Natural Language
  Learning}, pages 1157--1168.

\bibitem[{Marcheggiani and Titov(2017)}]{marcheggiani-titov-2017-encoding}
Diego Marcheggiani and Ivan Titov. 2017.
\newblock \href {https://doi.org/10.18653/v1/D17-1159} {Encoding sentences with
  graph convolutional networks for semantic role labeling}.
\newblock In \emph{Proceedings of the 2017 Conference on Empirical Methods in
  Natural Language Processing}, pages 1506--1515, Copenhagen, Denmark.
  Association for Computational Linguistics.

\bibitem[{Mostafazadeh et~al.(2016)Mostafazadeh, Chambers, He, Parikh, Batra,
  Vanderwende, Kohli, and Allen}]{mostafazadeh-etal-2016-corpus}
Nasrin Mostafazadeh, Nathanael Chambers, Xiaodong He, Devi Parikh, Dhruv Batra,
  Lucy Vanderwende, Pushmeet Kohli, and James Allen. 2016.
\newblock \href {https://doi.org/10.18653/v1/N16-1098} {A corpus and cloze
  evaluation for deeper understanding of commonsense stories}.
\newblock In \emph{Proceedings of the 2016 Conference of the North {A}merican
  Chapter of the Association for Computational Linguistics: Human Language
  Technologies}, pages 839--849, San Diego, California. Association for
  Computational Linguistics.

\bibitem[{Mostafazadeh et~al.(2017)Mostafazadeh, Roth, Louis, Chambers, and
  Allen}]{mostafazadeh-etal-2017-lsdsem}
Nasrin Mostafazadeh, Michael Roth, Annie Louis, Nathanael Chambers, and James
  Allen. 2017.
\newblock \href {https://doi.org/10.18653/v1/W17-0906} {{LSDS}em 2017 shared
  task: The story cloze test}.
\newblock In \emph{Proceedings of the 2nd Workshop on Linking Models of
  Lexical, Sentential and Discourse-level Semantics}, pages 46--51, Valencia,
  Spain. Association for Computational Linguistics.

\bibitem[{Nallapati et~al.(2017)Nallapati, Zhai, and
  Zhou}]{nallapati2017summarunner}
Ramesh Nallapati, Feifei Zhai, and Bowen Zhou. 2017.
\newblock Summarunner: A recurrent neural network based sequence model for
  extractive summarization of documents.
\newblock In \emph{Proceedings of the AAAI Conference on Artificial
  Intelligence}, volume~31.

\bibitem[{Pour et~al.(2020)Pour, Razavi, and Faili}]{pour2020new}
Melika~Golestani Pour, Seyedeh~Zahra Razavi, and Heshaam Faili. 2020.
\newblock A new sentence ordering method using bert pretrained model.
\newblock In \emph{2020 11th International Conference on Information and
  Knowledge Technology (IKT)}, pages 132--138. IEEE.

\bibitem[{Prabhumoye et~al.(2020)Prabhumoye, Salakhutdinov, and
  Black}]{prabhumoye-etal-2020-topological}
Shrimai Prabhumoye, Ruslan Salakhutdinov, and Alan~W Black. 2020.
\newblock \href {https://doi.org/10.18653/v1/2020.acl-main.248} {Topological
  sort for sentence ordering}.
\newblock In \emph{Proceedings of the 58th Annual Meeting of the Association
  for Computational Linguistics}, pages 2783--2792, Online. Association for
  Computational Linguistics.

\bibitem[{Rahutomo et~al.(2012)Rahutomo, Kitasuka, and
  Aritsugi}]{rahutomo2012semantic}
Faisal Rahutomo, Teruaki Kitasuka, and Masayoshi Aritsugi. 2012.
\newblock Semantic cosine similarity.
\newblock In \emph{The 7th International Student Conference on Advanced Science
  and Technology ICAST}, volume~4, page~1.

\bibitem[{Reiter and Dale(1997)}]{reiter1997building}
Ehud Reiter and Robert Dale. 1997.
\newblock Building applied natural language generation systems.
\newblock \emph{Natural Language Engineering}, 3(1):57--87.

\bibitem[{Song et~al.(2018)Song, Zhang, Wang, and Gildea}]{song-etal-2018-n}
Linfeng Song, Yue Zhang, Zhiguo Wang, and Daniel Gildea. 2018.
\newblock \href {https://doi.org/10.18653/v1/D18-1246} {N-ary relation
  extraction using graph-state {LSTM}}.
\newblock In \emph{Proceedings of the 2018 Conference on Empirical Methods in
  Natural Language Processing}, pages 2226--2235, Brussels, Belgium.
  Association for Computational Linguistics.

\bibitem[{Sutskever et~al.(2014)Sutskever, Vinyals, and
  Le}]{10.5555/2969033.2969173}
Ilya Sutskever, Oriol Vinyals, and Quoc~V. Le. 2014.
\newblock Sequence to sequence learning with neural networks.
\newblock In \emph{Proceedings of the 27th International Conference on Neural
  Information Processing Systems - Volume 2}, NIPS'14, page 3104–3112,
  Cambridge, MA, USA. MIT Press.

\bibitem[{Wang and Kuo(2020)}]{wang2020sbert}
Bin Wang and C-C~Jay Kuo. 2020.
\newblock Sbert-wk: A sentence embedding method by dissecting bert-based word
  models.
\newblock \emph{IEEE/ACM Transactions on Audio, Speech, and Language
  Processing}, 28:2146--2157.

\bibitem[{Wang and Wan(2019)}]{wang2019hierarchical}
Tianming Wang and Xiaojun Wan. 2019.
\newblock Hierarchical attention networks for sentence ordering.
\newblock In \emph{Proceedings of the AAAI Conference on Artificial
  Intelligence}, volume~33, pages 7184--7191.

\bibitem[{Wang et~al.(2018)Wang, Lv, Lan, and
  Zhang}]{wang-etal-2018-cross-lingual}
Zhichun Wang, Qingsong Lv, Xiaohan Lan, and Yu~Zhang. 2018.
\newblock \href {https://doi.org/10.18653/v1/D18-1032} {Cross-lingual knowledge
  graph alignment via graph convolutional networks}.
\newblock In \emph{Proceedings of the 2018 Conference on Empirical Methods in
  Natural Language Processing}, pages 349--357, Brussels, Belgium. Association
  for Computational Linguistics.

\bibitem[{Yanase et~al.(2015)Yanase, Miyoshi, Yanai, Sato, Iwayama, Niwa,
  Reisert, and Inui}]{yanase2015learning}
Toshihiko Yanase, Toshinori Miyoshi, Kohsuke Yanai, Misa Sato, Makoto Iwayama,
  Yoshiki Niwa, Paul Reisert, and Kentaro Inui. 2015.
\newblock Learning sentence ordering for opinion generation of debate.
\newblock In \emph{Proceedings of the 2nd Workshop on Argumentation Mining},
  pages 94--103.

\bibitem[{Yasunaga et~al.(2017)Yasunaga, Zhang, Meelu, Pareek, Srinivasan, and
  Radev}]{yasunaga-etal-2017-graph}
Michihiro Yasunaga, Rui Zhang, Kshitijh Meelu, Ayush Pareek, Krishnan
  Srinivasan, and Dragomir Radev. 2017.
\newblock \href {https://doi.org/10.18653/v1/K17-1045} {Graph-based neural
  multi-document summarization}.
\newblock In \emph{Proceedings of the 21st Conference on Computational Natural
  Language Learning ({C}o{NLL} 2017)}, pages 452--462, Vancouver, Canada.
  Association for Computational Linguistics.

\bibitem[{Yin et~al.(2020)Yin, Meng, Su, Ge, Song, Zhou, and
  Luo}]{yin2020enhancing}
Yongjing Yin, Fandong Meng, Jinsong Su, Yubin Ge, Lingeng Song, Jie Zhou, and
  Jiebo Luo. 2020.
\newblock Enhancing pointer network for sentence ordering with pairwise
  ordering predictions.
\newblock In \emph{Proceedings of the AAAI Conference on Artificial
  Intelligence}, volume~34, pages 9482--9489.

\bibitem[{Yin et~al.(2019)Yin, Song, Su, Zeng, Zhou, and Luo}]{ijcai2019-0748}
Yongjing Yin, Linfeng Song, Jinsong Su, Jiali Zeng, Chulun Zhou, and Jiebo Luo.
  2019.
\newblock \href {https://doi.org/10.24963/ijcai.2019/748} {Graph-based neural
  sentence ordering}.
\newblock In \emph{Proceedings of the Twenty-Eighth International Joint
  Conference on Artificial Intelligence, {IJCAI-19}}, pages 5387--5393.
  International Joint Conferences on Artificial Intelligence Organization.

\bibitem[{Yu et~al.(2018)Yu, Dohan, Luong, Zhao, Chen, Norouzi, and
  Le}]{yu2018qanet}
Adams~Wei Yu, David Dohan, Minh-Thang Luong, Rui Zhao, Kai Chen, Mohammad
  Norouzi, and Quoc~V Le. 2018.
\newblock Qanet: Combining local convolution with global self-attention for
  reading comprehension.
\newblock \emph{arXiv preprint arXiv:1804.09541}.

\bibitem[{Zeng et~al.(2018)Zeng, Li, Wang, Beauchamp, Shugars, and
  Wong}]{zeng2018microblog}
Xingshan Zeng, Jing Li, Lu~Wang, Nicholas Beauchamp, Sarah Shugars, and Kam-Fai
  Wong. 2018.
\newblock Microblog conversation recommendation via joint modeling of topics
  and discourse.
\newblock In \emph{Proceedings of the 2018 Conference of the North American
  Chapter of the Association for Computational Linguistics: Human Language
  Technologies, Volume 1 (Long Papers)}, pages 375--385.

\bibitem[{Zhang et~al.(2018)Zhang, Liu, and Song}]{zhang-etal-2018-sentence}
Yue Zhang, Qi~Liu, and Linfeng Song. 2018.
\newblock \href {https://doi.org/10.18653/v1/P18-1030} {Sentence-state {LSTM}
  for text representation}.
\newblock In \emph{Proceedings of the 56th Annual Meeting of the Association
  for Computational Linguistics (Volume 1: Long Papers)}, pages 317--327,
  Melbourne, Australia. Association for Computational Linguistics.

\bibitem[{Zhu et~al.(2020)Zhu, Song, Dou, Nie, and
  Zhou}]{zhu-etal-2020-scriptwriter}
Yutao Zhu, Ruihua Song, Zhicheng Dou, Jian-Yun Nie, and Jin Zhou. 2020.
\newblock \href {https://doi.org/10.18653/v1/2020.acl-main.765}
  {{S}cript{W}riter: Narrative-guided script generation}.
\newblock In \emph{Proceedings of the 58th Annual Meeting of the Association
  for Computational Linguistics}, pages 8647--8657, Online. Association for
  Computational Linguistics.

\end{thebibliography}
\bibliographystyle{acl_natbib}




\end{document}